\documentclass[review]{elsarticle}

\usepackage{hyperref}

\journal{Neurocomputing}

\usepackage{amsmath}
\usepackage{mathtools}
\usepackage{xcolor}
\usepackage{multirow}  
\usepackage{arydshln}  
\usepackage{soul}

\usepackage{algorithm,algpseudocode}
\usepackage{amsmath}
\usepackage{amsfonts}
\usepackage{amssymb}

\usepackage{wrapfig}

\newcommand{\etal}{\textit{et al.}}

\begin{document}

\begin{frontmatter}

\title{PseudoBound: Limiting the Anomaly Reconstruction Capability of One-Class Classifiers Using Pseudo Anomalies}

\author{Marcella Astrid$^{1,2}$, Muhammad Zaigham Zaheer$^{3}$, Seung-Ik Lee$^{1,2}$ \corref{mycorrespondingauthor}}
\address{$^{1}$University of Science and Technology, Daejeon, South Korea}
\address{$^{2}$Electronics and Telecommunications Research Institute, Daejeon, South Korea}
\address{$^{3}$Mohamed bin Zayed University of Artificial Intelligence, Abu Dhabi, United Arab Emirates}
\cortext[mycorrespondingauthor]{Corresponding author (Email: the\_silee@etri.re.kr)}




\begin{abstract}
   Due to the rarity of anomalous events, video anomaly detection is typically approached as one-class classification (OCC) problem. 
   Typically in OCC, an autoencoder (AE) is trained
   to reconstruct the normal only training data with the expectation that, in test time, it can poorly reconstruct the anomalous data. 
   However, previous studies have shown that, even trained with only normal data, AEs can often reconstruct anomalous data as well, resulting in a decreased performance. To mitigate this problem, we propose to limit the anomaly reconstruction capability of AEs by incorporating pseudo anomalies during the training of an AE.
   Extensive experiments using five types of pseudo anomalies show the robustness of our training mechanism towards any kind of pseudo anomaly. Moreover, we demonstrate the effectiveness of our proposed pseudo anomaly based training approach against several existing state-of-the-art (SOTA) methods on three benchmark video anomaly datasets, outperforming all the other reconstruction-based approaches in two datasets and showing the second best performance in the other dataset.
\end{abstract}

\begin{keyword}
anomaly detection\sep one class classification\sep pseudo anomaly
\end{keyword}

\end{frontmatter}

\section{Introduction}
\label{sec:introduction}

In recent years, artificial neural network/deep learning has proven its capability in a wide-range of applications, such as image classification \cite{yu2019hierarchical,wei2022generating}, human action recognition \cite{bai2022skeleton,nida2022video}, pose estimation \cite{hong2015multimodal,hong2018multimodal}, and graph classification \cite{ma2021aegcn,tran2018learning}. One of interesting application covered in this paper is anomalous event detection in videos. 

The anomaly detection problem has lately attracted significant attention but is extremely difficult due to its challenging nature, i.e., difficulties in collecting anomalous events as they rarely happen, especially in the real-life situations such as in surveillance videos. Moreover, the variety of anomalous events is unlimited, from riding bicycles on pedestrian pathways, explosion, earthquake, or gorilla attack, to any events that have never happened before. Despite the problem, several methods utilize both normal and anomalous data to train an anomalous event detection model, either with complete labels \cite{munawar2017limiting, yamanaka2019autoencoding} or weak labels \cite{sultani2018real,zaheer2020claws,zhong2019graph}. However, utilizing limited variety of anomalous events for training may also make the trained model overfit to the anomalous events provided in the training data. Therefore, anomaly detection models often rely only on normal data for training, making it a one-class classification (OCC) problem \cite{chang2020clustering,gong2019memorizing,zaheer2020old,park2021anomaly}. In this setting, anything deviating from the normal patterns in the training data is defined as anomalous.

\begin{figure}
\begin{center}
\includegraphics[width=\linewidth]{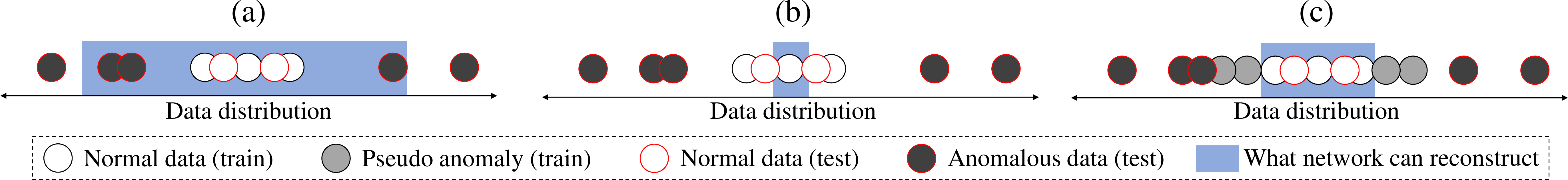}
\end{center}
\caption{An illustration of the reconstruction capability of the three variants of AE: (a) a conventional AE in which the reconstruction boundary can be anywhere as long as it includes the normal training data, (b) an AE with limited reconstruction capability so the normal data reconstruction may also be limited, and (c) an AE trained using pseudo anomalies to have more appropriate reconstruction boundary between normal and anomalous data.}
\label{fig:datadistribution}
\end{figure}

One typical way to handle the OCC problem in anomaly detection is by using a deep autoencoder (AE) to encode the normalcy information into its latent space 
\cite{luo2017revisit,hasan2016learning,zhao2017spatio,luo2017remembering}. To accomplish such objective, the model is trained to reconstruct normal data with the expectation that, at test time, it may poorly reconstruct anomalous cases while well reconstructing the normal cases.
However, AEs can also often well reconstruct anomalous examples (see the baseline performance in Fig. \ref{fig:qualitative}(a), (j), \& (r)). Similar observations have also been made by several previous works \cite{gong2019memorizing,zaheer2020old,zong2018deep,munawar2017limiting}. In such cases, the reconstructions between normal and anomalous data may not be distinguishable enough to successfully pin-point the anomalies. The phenomenon can be explained as in Fig. \ref{fig:datadistribution}(a).

There may be several ways to limit the reconstruction capability of an AE on anomalous data.
One naive approach could be to limit the network capacity itself by reducing the learnable parameters. However, this may also result in distorted normal data reconstructions, reducing the overall anomaly detection performance.
Another approach recently proposed by some researchers \cite{gong2019memorizing,park2020learning} is to add a memory mechanism between the encoder and the decoder of an AE to store normal features learned from training data. 
The AE is then limited to use the memorized normal definitions to reconstruct the inputs, which consequently reduces its ability to reconstruct anomalous data.
For instance, Fig. 6 of \cite{gong2019memorizing} and Fig. \ref{fig:qualitative}(i) show that 
a memory-based network can successfully distort more anomalous regions compared to the baseline. However, such a method relies on the memory size and too small memory may also limit its reconstruction capability on normal data. 
As a result, it may also be observed in Fig. 6 of \cite{gong2019memorizing} and Fig. \ref{fig:qualitative}(i) that some of the normal regions are distorted as well, which can limit the discrimination capability of the network. This phenomenon is illustrated in Fig. \ref{fig:datadistribution}(b). 

In order for an AE to learn more appropriate boundary between normal and anomalous data, we propose a more direct way to limit the reconstruction of abnormal regions by training an AE with the assistance of pseudo anomalies. The pseudo anomalies are generated to simulate data deviating from the normal data distribution.
The AE is then trained to minimize reconstruction loss for normal data while maximizing reconstruction loss for pseudo anomalous data. With this training setting, the AE is encouraged to limit its reconstruction capability only on out-of-distribution data, as illustrated in Fig. \ref{fig:datadistribution}(c).
We present evaluation of the proposed approach on three benchmark video anomaly detection datasets including Ped2 \cite{li2013anomaly}, Avenue \cite{lu2013abnormal}, and ShanghaiTech \cite{luo2017revisit}. Moreover, we provide detailed discussion on the impacts of several different types of pseudo anomalies on the training of an AE and lay out some insights on the design choices of utilizing pseudo anomalies successfully for the OCC.

A previous version of this work was presented in the International Conference on Computer Vision (ICCV) Workshop 2021 as ``Synthetic Temporal Anomaly Guided End-to-End Video Anomaly Detection" \cite{astrid2021synthetic} . Compared to the preceding preliminary version that uses only skipping frames as pseudo anomalies, in this conclusive work, we propose and extensively explore more types of pseudo anomalies to demonstrate the generic applicability of our training setup. The types of pseudo anomalies we experiment in this paper include skip frames, noise, patch, repeat frames, and fusion. Moreover, we extend the evaluations by further exploring the hyperparameters to observe the robustness of our method in different settings.

In summary, the contributions of our work are as follows:
1) We introduce a novel generic one class classification method by using pseudo anomalies in an end-to-end fashion. Using the training mechanism, our model successfully demonstrates distinctive reconstruction quality between normal data and anomalous data. 
2) We extensively explore and propose five different augmentation-based techniques to synthesize pseudo anomalies without ever using the real anomalies. 
3) By comparing different types of pseudo anomaly, we provide in-depth discussions on design choices for generating pseudo anomalies. Such an extensive study on pseudo anomalies, to the best of our knowledge, has not been carried out in the existing literature before.
4) We extensively evaluate the robustness of each technique using a wide-range of hyperparameters values.
5) We demonstrate the importance of our method in improving the anomaly detection performance of a conventional AE on three challenging benchmark datasets. 

\section{Related Works}
\label{sec:relatedworks}
Autoencoder (AE) is well-known for learning data representations in its latent space by reconstructing input. There are several applications that utilize AE to extract features, e.g., image classification \cite{diallo2021deep,yin2022semi}, human pose recovery \cite{hong2015multimodal,hong2016hypergraph}, and graph classification \cite{ma2021aegcn,tran2018learning}. In addition to the reconstruction task, several approaches have been proposed to learn better feature representations, e.g. learning both local and global characteristics of the dataset \cite{zhang2018local}, regularization \cite{salah2011contractive,hong2016hypergraph}, and clustering \cite{diallo2021deep,sheng2022contrastive}. 
In one-class classification (OCC) problem, similar to \cite{luo2017revisit,gong2019memorizing,hasan2016learning,zhao2017spatio,luo2017remembering}, we employ the reconstruction-based AE to learn the normal data representations using only normal training data. Nevertheless, our methods can also be extended to the aforementioned feature representation learning techniques.


As the AE is trained using only normal data, at test time, it is expected to poorly reconstruct anomalous data, corresponding to high anomaly scores. 
However, as observed by several researchers \cite{gong2019memorizing,zaheer2020old,zong2018deep,munawar2017limiting},  reconstruction-based AEs can often reconstruct anomalous data as well, reducing their discriminative capabilities.
Several researchers have attempted to limit the reconstruction capability of an AE on anomalous data. For example, memory-based networks \cite{gong2019memorizing,park2020learning} reconstruct the input using normalcy definitions saved inside the memory mechanism located between the encoder and the decoder of an AE.
However, such networks may also constrain the normal data reconstruction capability. 
In contrast, we try to limit the reconstruction capability of an AE on anomalous data by learning to maximize reconstruction loss on pseudo anomalies. 

Besides reconstruction-based AEs, several researchers adopt different strategies for anomaly detection. One idea is to detect only object-related anomalies by utilizing object detectors 
\cite{ionescu2019object,doshi2020any,sun2020scene,yu2020cloze,georgescu2021background}.  
Another idea is to utilize components such as optical flow \cite{liu2018future,lee2019bman,nguyen2019anomaly,wu2019deep,ravanbakhsh2017abnormal} or RGB-difference \cite{chang2020clustering} in order to add motion information to the anomaly detectors. Similar in essence, prediction-based approaches \cite{liu2018future,park2020learning,lu2019future,lu2020few,dong2020dual} attempt to implicitly incorporate the motion information by predicting future frames given the past few consecutive frames as inputs. 
Compared to the reconstruction-based approaches, prediction methods try to reconstruct future frames rather than the input frames.
In order to improve the reconstruction quality of AEs, some researchers also propose to incorporate adversarial training 
\cite{liu2018future,lee2019bman,vu2019robust,ji2020tam,lee2018stan}. 
Since these methods are not characterized as reconstruction-based, our approach is significantly different. However, the goal is common, i.e. detecting anomalies.

There has been few recent attempts toward pseudo anomaly generation for one-class classifiers.
OGNet \cite{zaheer2020old} and G2D \cite{pourreza2021g2d} propose to early stop an adversarially trained generator and use it to generate fake anomaly data, which is then utilized together with the normal data to train a binary classifier.
Additionally, OGNet fuses two images then passes it to a fully-trained generator to produce another type of fake anomalous example.
In contrast, our pseudo anomaly synthesizer does not require any pre-trained network to generate the pseudo anomalous samples. More recently, Georgescu \etal {} \cite{georgescu2021background} propose time magnification and replacing with separate dataset to generate pseudo abnormal objects from the detected objects. In contrast, our model does not rely on any object detectors and generates pseudo anomalies in frame-level instead of object-level. 
Another recent work, Astrid \etal {}  \cite{astrid2021learning} also utilize skip frame based and patch based pseudo anomalies. The idea is to train the network to reconstruct normal regardless of the input, i.e., normal or pseudo anomaly. On the other hand, in this work, the training setup is to minimize the reconstruction loss in the case of normal data and maximize it in the case of pseudo anomaly data. Since \cite{astrid2021learning} and this work are closely related, i.e., both are reconstruction-based methods that use pseudo anomalies, we also provide detailed comparisons in Section \ref{subsubsec:comparewithlntra}.

In order to improve the discrimination capability of an anomaly detection model, several researchers \cite{munawar2017limiting, yamanaka2019autoencoding} utilize real anomalies during training time, deviating from the OCC definition. 
We also acknowledge several recently-introduced weakly supervised methods which use video-level anomaly/normal annotations for training \cite{sultani2018real,zaheer2020claws,zhong2019graph}.
However, collecting sufficient number of diverse anomalous examples can be cumbersome as such events rarely occur. Our method, on the other hand, follows the conventional OCC protocol by using normal data and pseudo anomaly examples synthesized from normal data to train our model.

Our approach can also be seen as a type of data augmentation technique, typically used in image classification \cite{bengio2011deep,krizhevsky2012imagenet,yun2019cutmix,lee2020smoothmix}, which manipulates training data to increase variety. In image classification, the class labels for the augmented data are obtained from the original classes in the train set. In contrast, the augmentation in our method produces a new class, i.e. anomaly, which is not a part of the original classes in the train set. 

\begin{figure}
\begin{center}
\includegraphics[width=\linewidth]{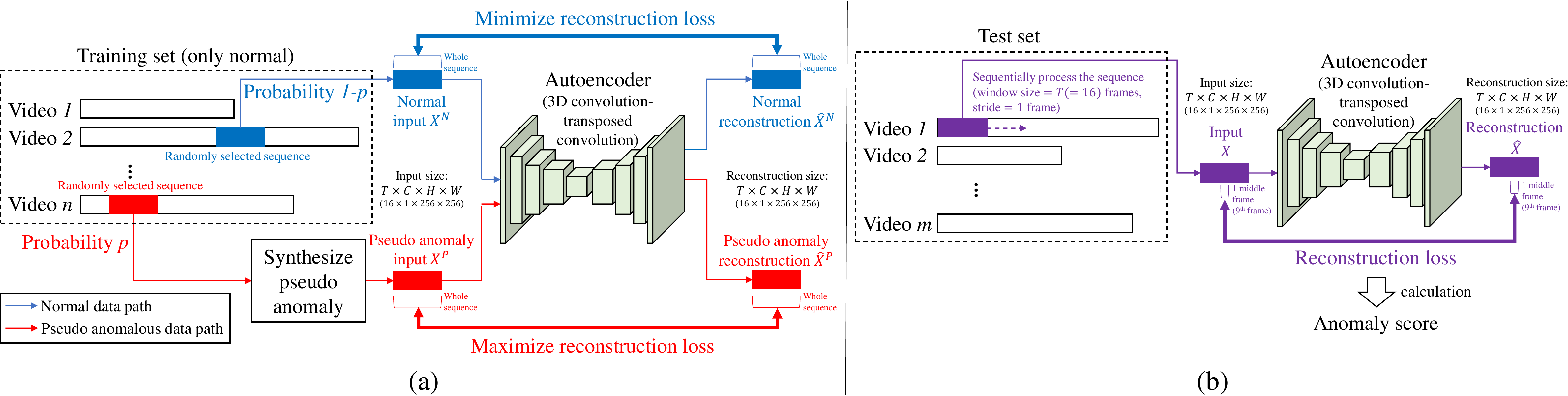}
\end{center}
\caption{
Our model takes input of size $T$ frames $\times$ channel $C \times$ height $H \times$ width $W$ and outputs reconstruction of the input of the same size. 
(a) During training, with a probability $1-p$, we randomly select a sequence of size $T$ frames as a normal input $X^N$ from training videos which are only normal, and transform it into reconstruction $\hat{X}^N$. The autoencoder (AE) is then trained to minimize reconstruction loss of the whole sequence of $T$ frames. With a probability $p$, using a selected method of synthesizing pseudo anomalies, a pseudo anomalous input $X^P$ is synthesized from normal data. $X^P$ is then processed by the AE to produce reconstruction $\hat{X}^P$. In this pseudo anomaly case, the AE is trained to maximize the reconstruction loss of the whole input sequence. (b) During testing, we take $T$ frames in order from test set then give it as input to the trained AE. The anomaly score is finally calculated from the reconstruction loss between the middle frame of $T$ frames input and middle frame of the $T$ frames reconstruction. 
   }
\label{fig:overalltraintest}
\end{figure}

Besides surveillance in videos, as also reviewed in \mbox{\cite{pang2021deep,pang2021toward}}, there are several other applications of anomaly detection models, for example network \mbox{\cite{samriya2022network,gong2019memorizing}}, blockchain \mbox{\cite{fan2021spsd,hu2021transaction}}, sensors \mbox{\cite{park2021anomaly,li2021entropy}}, general images \mbox{\cite{cai2022perturbation,salehi2021arae}}, and medical images \mbox{\cite{cai2022perturbation,salehi2021arae}}, and industrial image data \mbox{\cite{murase2022algan,roth2022towards}}. The idea of pseudo anomalies can also be explored in these other applications, for example, ALGAN {\cite{murase2022algan}} utilizes a less-trained generator and generator with anomalous latent input to produce pseudo anomalies in industrial image application. Another example, PLAD {\cite{cai2022perturbation}} uses perturbed images as pseudo anomalies for image data. Several types of pseudo anomalies proposed in this work may also be applicable towards applications outside surveillance videos, which can be a possible future direction of our current research work.

\section{Methodology}

In this section, we present PseudoBound, our approach to limit the anomaly reconstruction capability of one-class classifiers using pseudo anomalies. The training mechanism using normal and pseudo anomalies of PseudoBound is illustrated in Fig. \ref{fig:overalltraintest}(a) which is further detailed with the pseudo anomaly types selection in Fig. \ref{fig:overall}. 
The corresponding discussion of the training mechanism is provided in Section \ref{subsec:trainingusingpseudoanomalies}. Section \ref{subsec:synthesizingpseudoanomalies} describes our method of synthesizing pseudo anomalies.
The test mechanism is illustrated in Fig. \ref{fig:overalltraintest}(b) and further described in Section \ref{subsec:anomalyscore}.

\subsection{Training using normal and pseudo anomalous data}
\label{subsec:trainingusingpseudoanomalies}

In order to capture robust representations from temporal and spatial domains, similar to \cite{gong2019memorizing,park2020learning,hasan2016learning,zhao2017spatio}, we set up our autoencoder (AE) model to take multi-frame input $X$ of size $T \times C \times H \times W$ and produce the reconstruction $\hat{X}$ of the same size, where $T$ , $C$, $H$, and $W$ are the number of frames, number of channels, height, and width of frames in the input sequence. Typically, the training is carried out by minimizing the reconstruction loss between $X$ and $\hat{X}$. Moreover, to stabilize the training, in this paper, we limit the range of $\hat{X}$ and $X$ values to $[-1, 1]$.

Conventionally, an AE is trained only on normal data with the expectation to produce low reconstruction loss on normal data and high reconstruction loss on anomalous data during test time. However, AEs can often ``generalize" too well and start reconstructing anomalous examples as well \cite{gong2019memorizing,zaheer2020old,zong2018deep,munawar2017limiting}. 
Therefore, we need a mechanism that enforces high reconstruction loss on anomalous data without using real anomalous data itself. 
To this end, we propose a training mechanism using pseudo anomalies $X^P$ in addition to the normal data $X^N$ as:
\begin{equation}
    X= 
\begin{cases}
    X^{N} & \text{with probability $1-p$,}\\
    X^{P} & \text{with probability $p$,}\\
\end{cases}
\label{eq:inputselection}
\end{equation}
where $p \in [0, 1]$ is the probability hyperparameter defining the amount of pseudo anomaly examples relative to the normal data.

\begin{figure}
\begin{center}
\includegraphics[width=\linewidth]{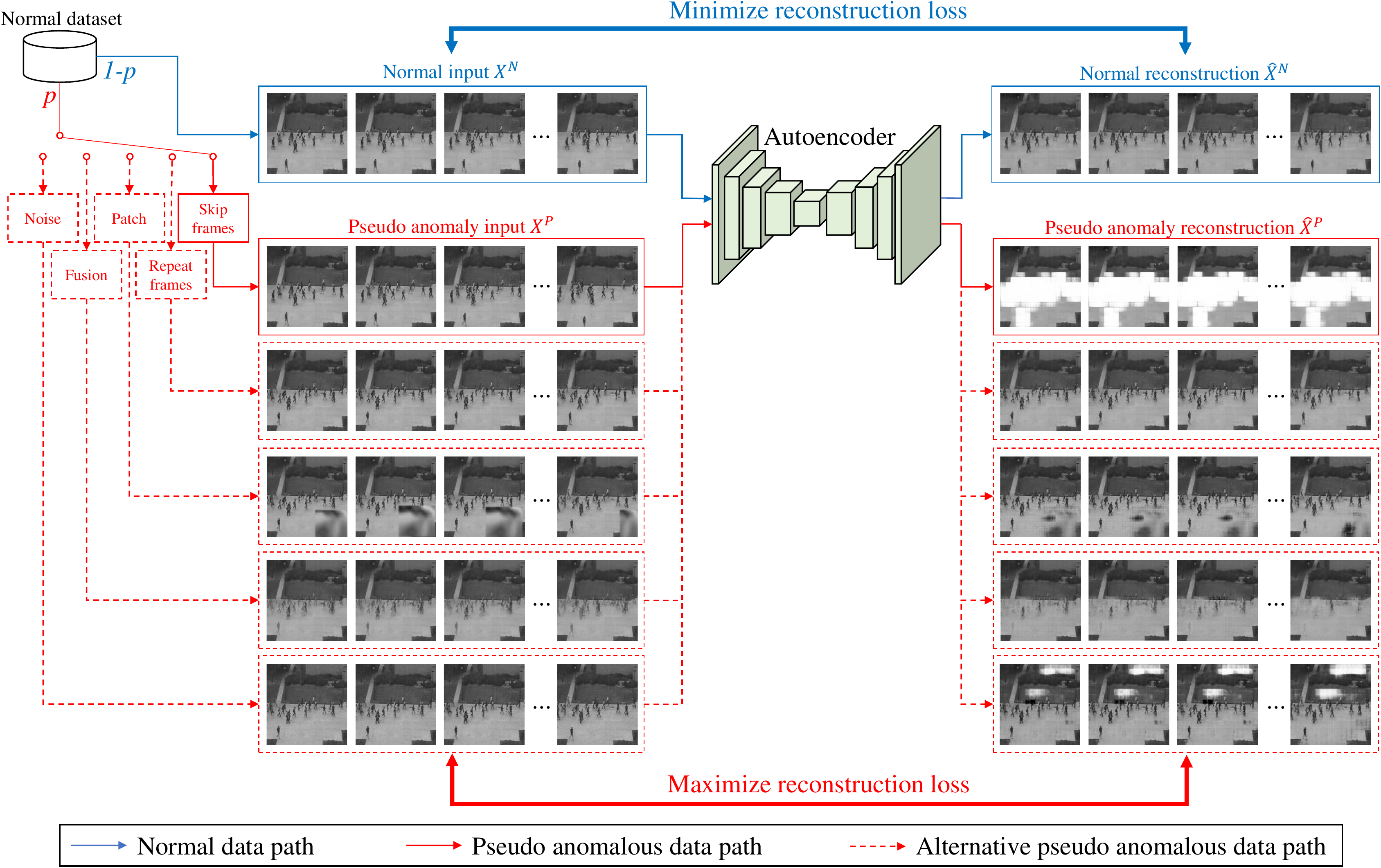}
\end{center}
\caption{
In order for an AE to reconstruct normal data well while distorting anomalous data reconstruction,
we train an AE using normal and pseudo anomalous data. Pseudo anomalous data is used with a probability $p$ whereas normal data with a probability $1-p$. In the case of normal data, the AE is trained to minimize reconstruction loss while maximizing it in the case of pseudo anomalous input. In this work, we propose five different methods to synthesize pseudo anomalies that can be used to train an efficient AE for the purpose of anomaly detection. 
   }
\label{fig:overall}
\end{figure}

In order to create a discriminative reconstruction between normal and pseudo anomalies, the AE is trained to minimize the reconstruction loss for normal data while maximizing it for pseudo anomaly examples. The overall loss of our proposed system is then defined as:
\begin{equation}
    L= 
\begin{cases}
    L^{N} & \text{if } X=X^{N} \text{,}\\
    L^{P} & \text{if } X=X^{P} \text{,}\\
\end{cases}
\label{eq:reconloss}
\end{equation}

where $L^N$ is the reconstruction loss for normal data as
\begin{equation}
    L^N= \frac{1}{T \times C \times H \times W}  \left \| \hat{X}^N - X^N  \right \|_{F}^{2} \text{,}
\label{eq:aereconloss}
\end{equation}
and $L^{P}$ is the reconstruction loss for pseudo anomaly data as:
\begin{equation}
    L^{P}= - \frac{1}{T \times C \times H \times W} \left \| \hat{X}^P - X^P  \right \|_{F}^{2}  \text{,}
\label{eq:pseudoreconloss}
\end{equation}
where $\left \| .  \right \|_{F}$ means Frobenius norm.
Note the negative sign in Eq. \eqref{eq:pseudoreconloss}, which is introduced to maximize the reconstruction loss of pseudo anomaly examples. As discussed in Section \ref{sec:experiments}, despite not using real anomalies, this training configuration helps in limiting the reconstruction capability of an AE on real anomalous inputs during test time.


For the normal data $X^N$, similar to the typical procedures in training a conventional AE \cite{hasan2016learning,park2020learning}, we randomly extract a consecutive sequence of normal frames $X^{N}$ from a randomly selected training video $V = \{I_1, I_2, ..., I_{l} \}$ as:
\begin{equation}
\begin{split}
    X^{N} & = (I_n, I_{n+1}, ..., I_{n+T-1}) \\ & = (I_{n+t})_{0 \leq t < T, \ n+T-1 \leq l} \text{, }
\label{eq:sequencenormal}
\end{split}
\end{equation} 
where $l$ and $n$ are video length and the starting frame index of $X^{N}$ in $V$, respectively. For the pseudo anomalous data $X^P$, we discuss further in Section \ref{subsec:synthesizingpseudoanomalies}.

The overall training configuration with a given pseudo anomaly type can be seen in Fig. \ref{fig:overalltraintest}(a). In this work, there are five different types of pseudo anomalies that can be selected, as shown in Fig. \ref{fig:overall}. In practice, the overall training process in an iteration can also be seen in Algorithm {\ref{algo:training}}. Lines 7-13 represent the input selection in Eq. {\eqref{eq:inputselection}}. The training loss based on the input selection in Eq. {\eqref{eq:reconloss}} can be seen in Lines 18-22.


\begin{algorithm}
\caption{Training process in an iteration.}
\label{algo:training}
\begin{algorithmic}[1]
\Require{a batch of normal data $X^N[1, \cdots, b]$, a batch of pseudo anomalous data $X^P[1, \cdots, b]$, autoencoder $\mathcal{AE}$, uniform distribution from zero to one $\mathcal{U}$, Adam optimizer $Adam$} 
\Ensure{batch size $b$, probability of pseudo anomaly $p$}
\State{$X^N \gets$ sample $b$ number of normal data}
\State{$X^P \gets$ sample $b$ number of pseudo anomalous data}
\State{$X \gets [\;]$}
\State{$pseudo\_anomaly\_status \gets [\;]$}
\For{$i = 1$ to $b$}                    
    \State {Sample $rand\_number \sim \mathcal{U}$ }
    \If{$rand\_number < p$}  \Comment{Eq. \eqref{eq:inputselection}}
        \State{$X \gets$ append $X^P[i]$}
        \State{$pseudo\_anomaly\_status \gets$ append $True$}
    \Else
        \State{$X \gets$ append $X^N[i]$}
        \State{$pseudo\_anomaly\_status \gets$ append $False$}
    \EndIf 
\EndFor
\State{$\hat{X} \gets \mathcal{AE}(X)$}
\State{$L\gets0$}
\For{$i \gets 1$ to $b$}
    \If{$pseudo\_anomaly\_status[i]$} 
        \State{$L \gets L - \frac{1}{T \times C \times H \times W} \left \| \hat{X}^P - X^P  \right \|_{F}^{2}$} \Comment{Eq. \eqref{eq:reconloss} and Eq. \eqref{eq:pseudoreconloss}}
    \Else
        \State{$L \gets L + \frac{1}{T \times C \times H \times W}  \left \| \hat{X}^N - X^N  \right \|_{F}^{2}$} \Comment{Eq. \eqref{eq:reconloss} and Eq. \eqref{eq:aereconloss}}
    \EndIf 
\EndFor
\State{$L \gets L / b$}
\State{$\mathcal{AE} \gets Adam(L, \mathcal{AE})$} \Comment{Update the AE}
\end{algorithmic}
\end{algorithm}

\subsection{Synthesizing pseudo anomalies}
\label{subsec:synthesizingpseudoanomalies}

We propose to generate pseudo anomalies by manipulating the normal training data to produce samples outside the normal data distribution.
Pseudo anomalies are not real anomalies, hence the name `pseudo'. Moreover, they are anomalous since their patterns do not exist in the training data that defines the normal data distribution. In this section, we introduce five different ways to construct pseudo anomalies in a simple manner, in which they do not require any pretrained network or any additional network architecture.

\subsubsection{Skip frames}

We design our skipping frames pseudo anomaly synthesizer based on the intuition that detecting fast or suddenly changing motions is significantly important and closely related to anomalous events.
For instance, people may run unusually if there are life-threatening situations nearby, such as natural disasters \cite{hu2010study,li2015parameter} or fires \cite{keating1982myth,elliott1993football}.
Moreover, robberies and fights may also involve sudden strong motions.
Some other examples may include over-speeding vehicles, riding bikes or vehicles on pedestrian lanes, etc. \cite{aultman1999toronto,sikka2019sharing,muthusamy2015review}. To mimic such anomalous movements, as shown in Fig. \ref{fig:skiprepeatframepseudoanomaly}(c), we propose to arbitrarily skip few frames for generating  pseudo anomalies:
\begin{equation}
\begin{split}
    X^P & = (I_n, I_{n+s}, ..., I_{n+(T-1)s}) \\ & = (I_{n+ts})_{0 \leq t < T, \ n+(T-1)s \leq l, \ s>1} \text{, }
\label{eq:sequenceskipframepseudo}
\end{split}
\end{equation}
where $s$ is a hyperparameter to control the number of frames we skip for generating pseudo anomaly examples. For generalization, we assign $s$ into a set of numbers, e.g. $s=\{2,3\}$ means $s$ is randomly selected as $2$ or $3$ each time a pseudo anomaly sequence is generated.

\subsubsection{Repeat frames}
\begin{figure}
\begin{center}
\includegraphics[width=\linewidth]{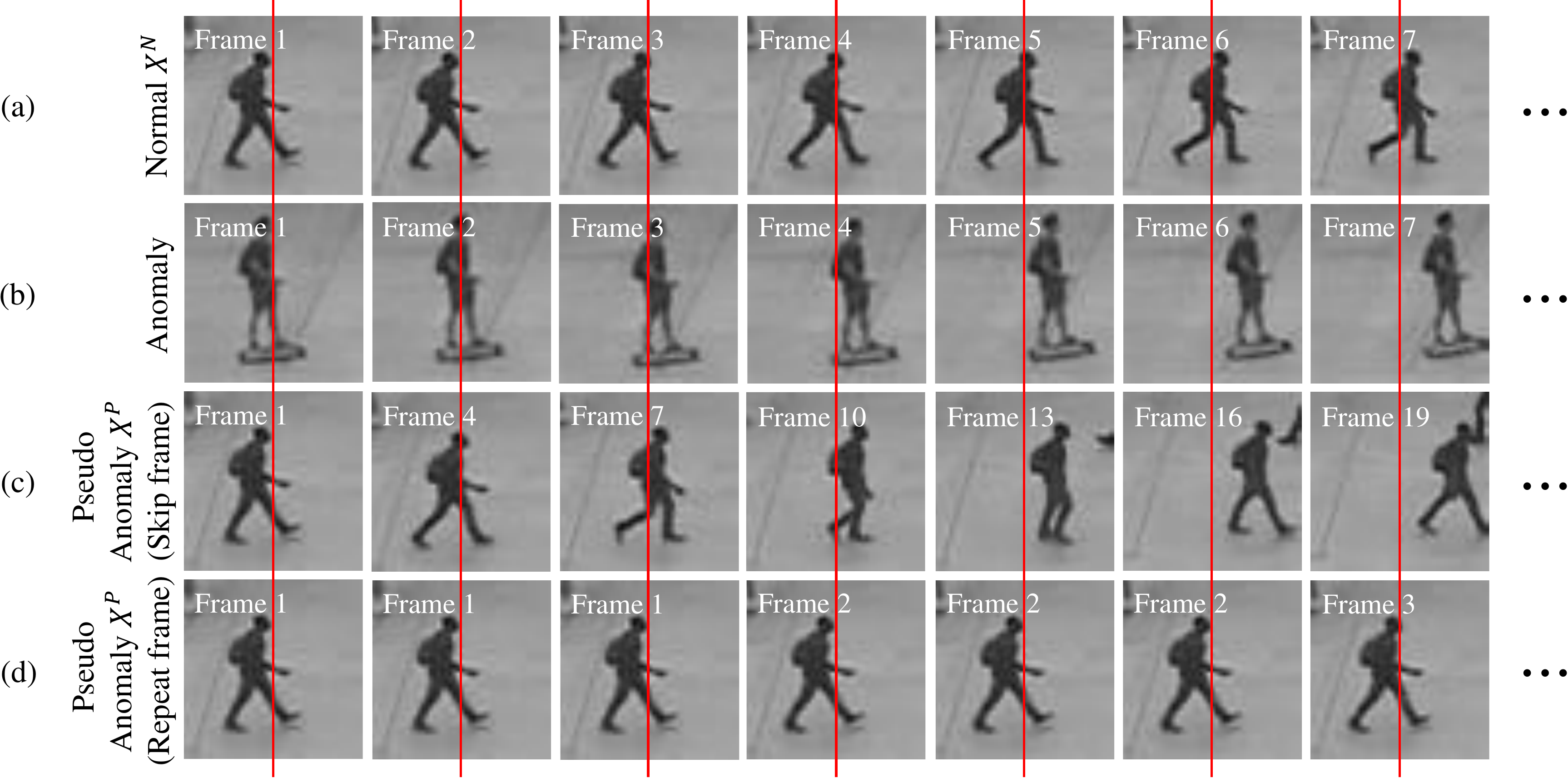}
\end{center}
\caption{Visualization of normal, real anomalous (test), skip frame pseudo anomalies (train), and repeat frame pseudo anomalies (train). Given red lines as references, (a) shows a normal movement in which the subject barely crosses the red line, (b) shows how the subject with anomalous movement completely crosses the reference line just in few frames, (c) shows how our skip frame pseudo anomalous frames with $s=3$  mimic the real anomalous movements by using normal data only, (d) shows how repeating frames with $r=3$ makes the subject moves slower than the normal sequence in which the object does not cross the red line.}
\label{fig:skiprepeatframepseudoanomaly}
\end{figure}

Pseudo anomalies created based on skip frames may be considered as utilizing the prior knowledge that many anomalies are related to fast movements. In order to further explore if our method is generic enough to overcome such assumption, we propose to experiment with the opposite of the prior.
To this end, we use repeating frames to mimic the kind of slow movements, which certainly do not appear in the normal training data. These slow motion pseudo anomalies are given as: 
\begin{equation}
\begin{split}
    X^P & = (I_n, ..., I_n, I_{n+1}, ..., I_{n+1}, ..., I_{n+\lfloor \frac{T-1}{r} \rfloor}, ..., I_{n+\lfloor \frac{T-1}{r} \rfloor}) \\ & = (I_{n+\lfloor \frac{t}{r} \rfloor})_{0 \leq t < T, \ n+\lfloor \frac{T-1}{r} \rfloor  \leq l, \ r>1} \text{, }
\label{eq:sequencerepeatframepseudo}
\end{split}
\end{equation}
where $r$ controls the number of repetition of a frame before moving to the next frame and $\lfloor \: \rfloor$ denotes the floor function. Similar to $s$ in the skip frames, we can also assign the $r$ values into a set of numbers. Visualization on how repeating frames can mimic slow movements can be seen in Fig. \ref{fig:skiprepeatframepseudoanomaly}(d).

\subsubsection{Patch}

In video anomaly detection, there are anomaly cases that are related to the appearance of some objects. Similar with motion related anomalies, this kind of appearance related anomalies are not available as well in the training data set in the OCC setting. In order to overcome this, we propose to make an image patch from an intruder dataset and paste it into each frames of an input sequence.

\begin{figure}
\begin{center}
\includegraphics[width=\linewidth]{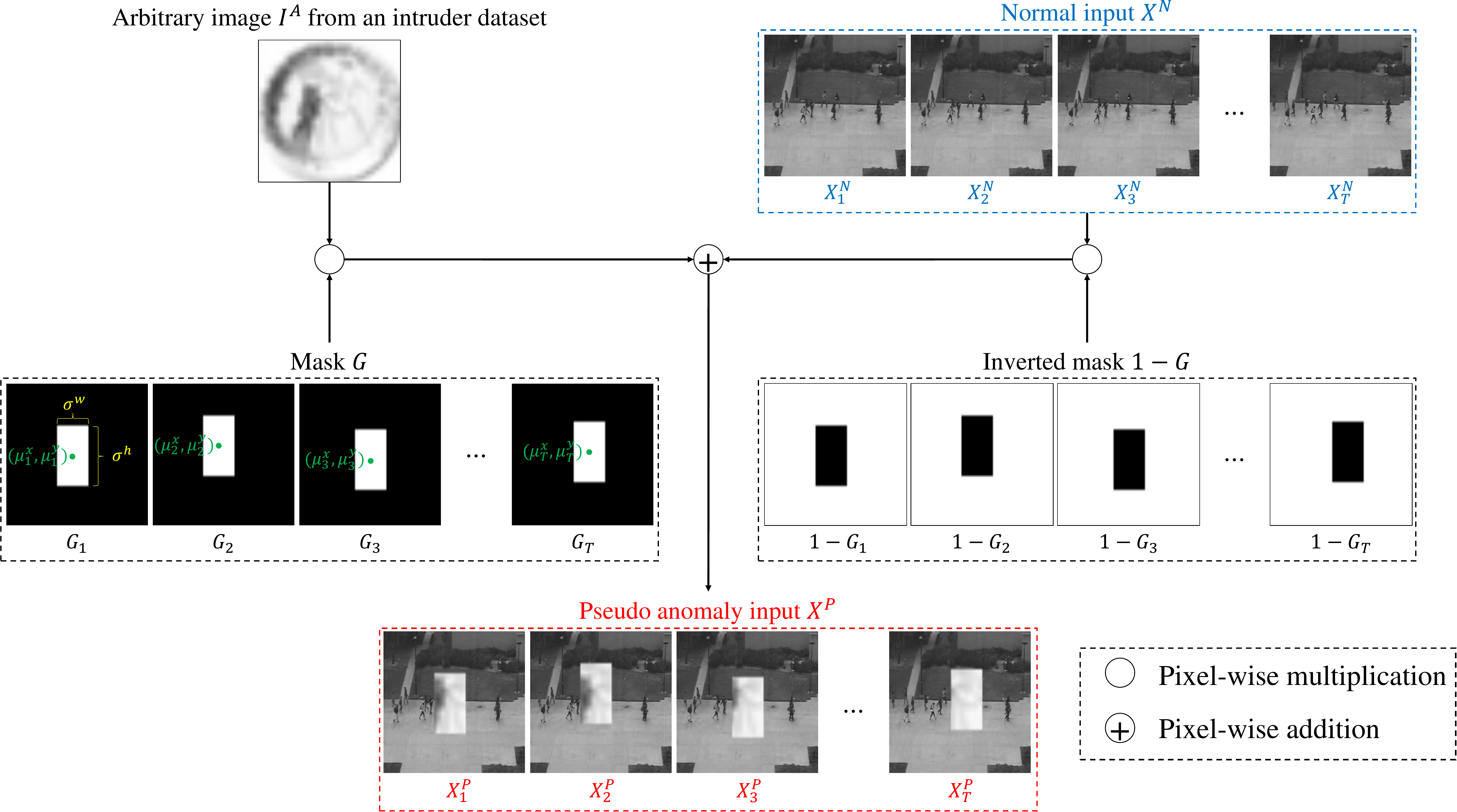}
\end{center}
\caption{Patch based pseudo anomaly construction by putting patch from the intruder dataset. In this example, SmoothMixS \cite{lee2020smoothmix} and CIFAR-100 \cite{krizhevsky2009learning} are used as the patching technique and intruder dataset, respectively.}
\label{fig:patchpseudoanomaly}
\end{figure}

To generate the $i$-th frame of the pseudo anomaly $X^{P}_i$ of input sequence, we combine an arbitrary image $I^A$ from an intruding dataset and the $i$-th normal frame $X^N_i$ of the sequence as follows:
\begin{equation}
    X^{P}_i = G_i \circ I^A + (1-G_i) \circ X^N_i \text{, }
\label{eq:patchpseudo}
\end{equation}
where $\circ$ is pixel-wise multiplication in the spatial dimension, $G_i$ is mask for the $i$-th frame of size $H \times W$, and $I^A$ is an arbitrary image taken from the intruding dataset and transformed to have the same size as $X^N_i$, i.e., $C \times H \times W$. The mask can be constructed using different patching techniques, such as SmoothMixS \cite{lee2020smoothmix}, SmoothMixC \cite{lee2020smoothmix}, and CutMix \cite{yun2019cutmix}. 
Furthermore, besides patching an image $I^A$ into all of the input frames, we can also put an arbitrary sequence taken from an intruder video dataset, where each frame $\{I^A_1, I^A_2, ..., I^A_T\}$ is correspondingly used to patch and create each pseudo anomalous frame $\{X^P_1, X^P_2, ..., X^P_T\}$.

\begin{figure}
\begin{center}
\includegraphics[width=0.8\linewidth]{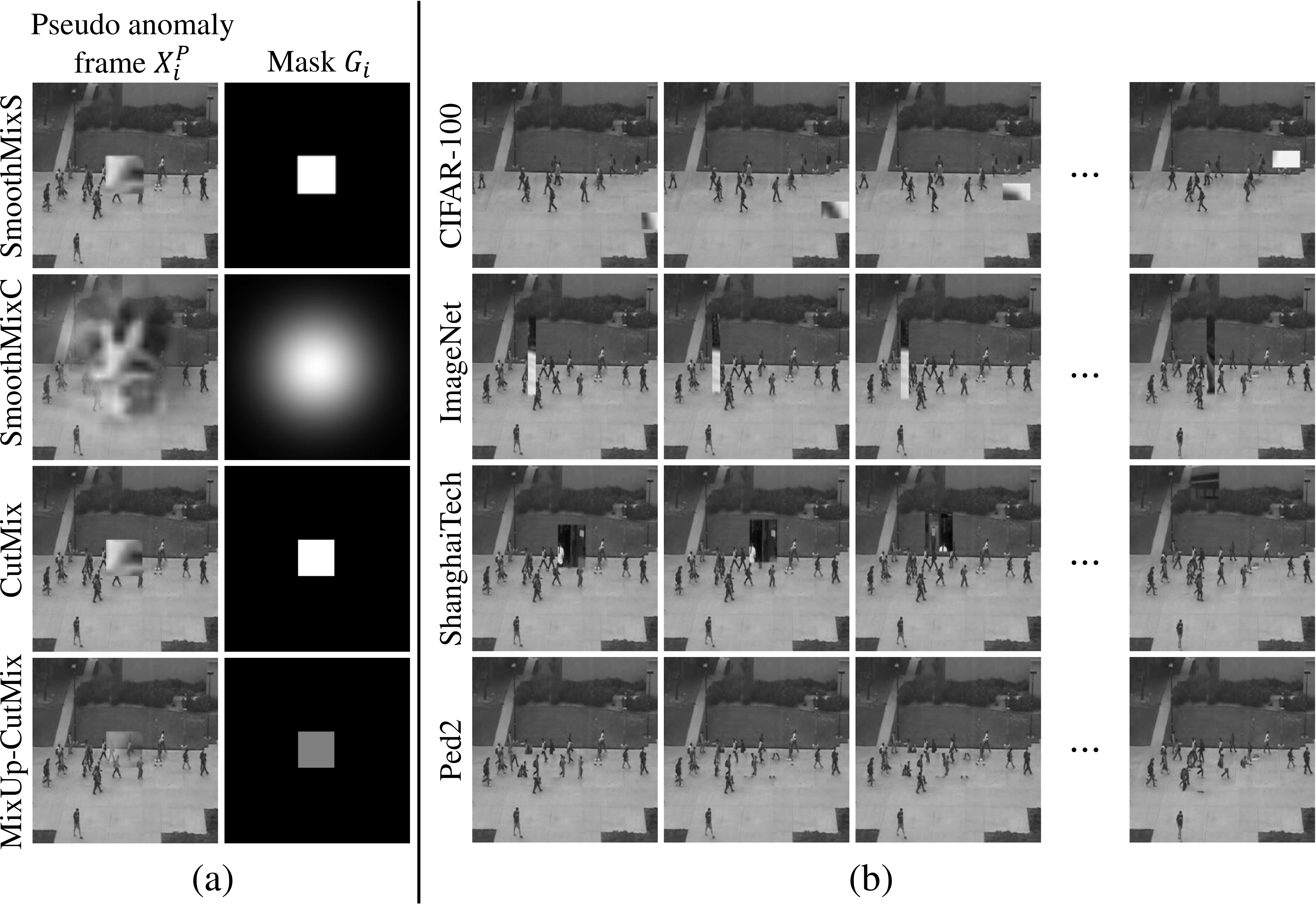}
\end{center}
\caption{Patch based pseudo anomalies in Ped2 using (a) various patching techniques and (b) intruder datasets. Every patch in (a) is set to $\mu^x = \mu^y = 127$ and $\sigma^w = \sigma^h = 50$. Pseudo anomalies in (b) are all constructed using SmoothMixS \cite{lee2020smoothmix}. }
\label{fig:variouspatchpseudoanomaly}
\end{figure}

Using a patching technique, each mask $G_i$ is constructed with parameters $(\mu^x_i, \mu^y_i)$ and $(\sigma^w, \sigma^h)$ as the center position and size of the patch, respectively. The position is randomly selected across the image space whereas the size is randomly selected within the range $[10, \alpha W]$ and $[10, \alpha H]$, where $\alpha$ is a hyperparameter controlling the maximum size of the patch relative to the frame size. Moreover, as anomalous objects in videos are usually moving, we also incorporate patch movements by changing the center position as:
\begin{equation}
    \mu_{i}^x = \mu_{i-1}^x + \Delta\mu_i^x \text{,} \;\;\;\;\;\;
    \mu_{i}^y = \mu_{i-1}^y + \Delta\mu_i^y \text{, }
\label{eq:movingpatch}
\end{equation}
where each $\Delta\mu_i^x$ and $\Delta\mu_i^y$ is randomly selected within the range of $[-\beta, \beta]$ and $\beta$ is a hyperparameter to adjust the maximum movement of the patch in terms of pixels.
Visualization of the patch based pseudo anomaly can be seen in Fig. \ref{fig:patchpseudoanomaly} and \ref{fig:variouspatchpseudoanomaly}.
Note that, using the training data set as the intruding data set, e.g., Ped2 patch masked into Ped2 normal frames, is also possible, which can result in pseudo anomaly objects, such as cropped pedestrians in Fig. \ref{fig:variouspatchpseudoanomaly}(b)-Ped2.



\subsubsection{Fusion}

\begin{figure}
\begin{center}
\includegraphics[width=\linewidth]{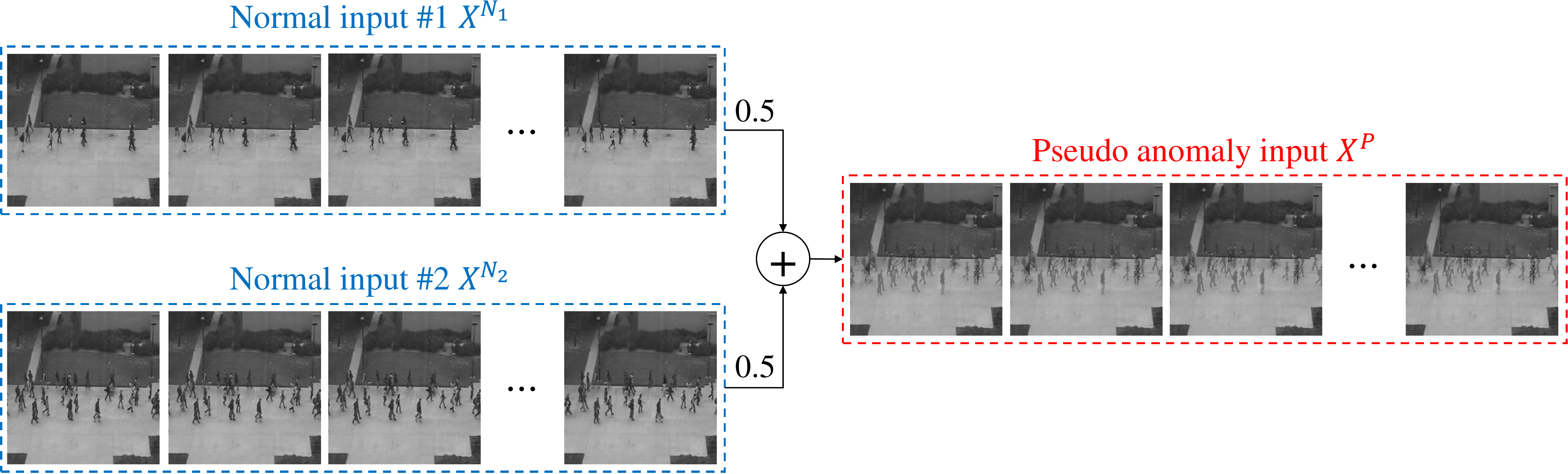}
\end{center}
\caption{Generation of pseudo anomalies by fusing two randomly selected normal sequences which produces shadows or unusual shapes that are not normal.}
\label{fig:averagepseudoanomaly}
\end{figure}

Inspired by OGNet \cite{zaheer2020old}, we generate a pseudo anomaly by combining two randomly selected sequences $X^{N_1}$ and $X^{N_2}$ taken from the normal training data as follows:
\begin{equation}
\begin{multlined}
    X^{P} = \frac{X^{N_1} + X^{N_2}}{2} \text{. }
\label{eq:sequenceaveragepseudo}
\end{multlined}
\end{equation}
This way, the resultant image $X^{P}$ may contain shadows or unusual shapes that are not present in the normal training data.
Fig. \ref{fig:averagepseudoanomaly} illustrates the pseudo anomaly generated by fusing two normal sequences.
It may also be noted that unlike OGNet \cite{zaheer2020old}, our synthesizer does not employ any pre-trained model thus making our method end-to-end trainable.

\subsubsection{Noise}

\begin{figure}
\begin{center}
\includegraphics[width=0.8\linewidth]{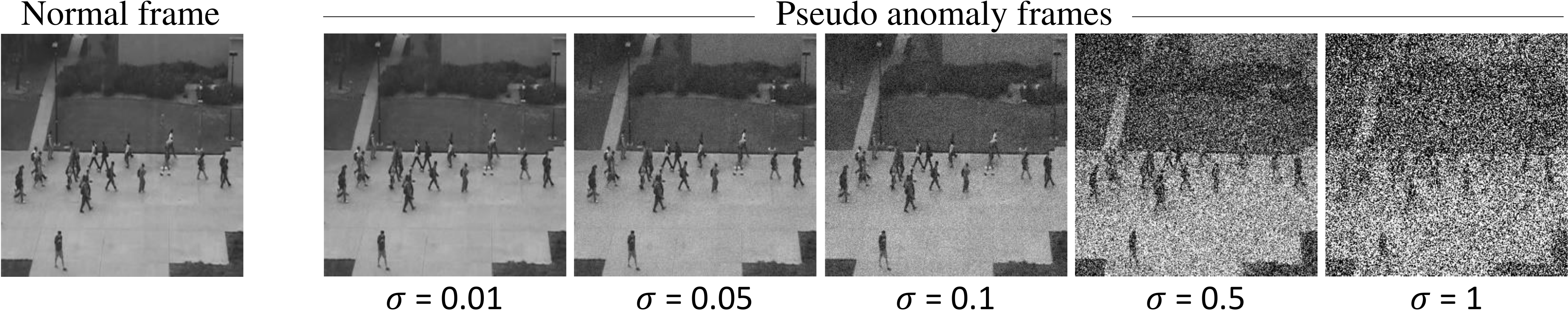}
\end{center}
\caption{Visualization of a normal frame and its respective pseudo anomaly frames constructed by adding Gaussian noise with different standard deviation $\sigma$ values.}
\label{fig:noisepseudoanomaly}
\end{figure}

Constructing pseudo anomalies with skipping frames requires the prior knowledge that anomalous behaviors are closely related to fast movements. Similarly, overlaying frames with patches or fusing two normal frames also requires prior knowledge on anomalous objects and appearances. On the contrary, adding noise as proposed here constructs pseudo anomalies without any prior knowledge by simply adding noise to the normal input:
\begin{equation}
    X^P = X^N + \mathcal{G}(0, \sigma) \text{, }
\label{eq:noisepseudo}
\end{equation}
where $\mathcal{G}(0, \sigma)$ is Gaussian noise with mean 0 and standard deviation $\sigma$. Fig. \ref{fig:noisepseudoanomaly} shows some examples of noise based pseudo anomalies constructed using different $\sigma$ values. 

\subsection{Anomaly score}
\label{subsec:anomalyscore}

Concurrent to previous approaches \cite{gong2019memorizing,park2020learning,liu2018future,hasan2016learning,zhao2017spatio,zaheer2020old}, we compute frame-level anomaly scores at test time. As Peak Signal to Noise Ratio (PSNR) is generally better metric for image quality compared to the reconstruction loss \cite{mathieu2015deep}, similar to \cite{park2020learning,liu2018future,dong2020dual}, we calculate PSNR $\mathcal{P}_t$ between an input frame and the reconstruction as:
\begin{equation}
    \mathcal{P}_t = 10 \text{ log}_{10}  \frac{M_{\hat{I}_t}^2}{\frac{1}{R} \left \| \hat{I}_t - I_t  \right \|_{F}^{2} } \text{,}
\label{eq:psnr}
\end{equation}
where $t$ is the frame index, $I_t$ is the $t$-th frame input, $\hat{I}_t$ is the reconstruction of $I_t$, $R$ is the total number of pixels in $\hat{I}_t$, and $M_{\hat{I}_t}$ is the maximum possible pixel value difference ($M_{\hat{I}_t} = 2$).

Then, following \cite{park2020learning,liu2018future,dong2020dual}, the PSNR value is normalized to the range of $[0,1]$ over all the frames in a test video $V_i$ as follows:
\begin{equation}
    \mathcal{Q}_t = \frac{\mathcal{P}_t - \min_t(\mathcal{P}_t)}{\max_t(\mathcal{P}_t)-\min_t(\mathcal{P}_t)} \text{,}
\label{eq:minmax}
\end{equation}
where $t$ is the frame index of $V_i$. A lower $Q_t$ represents higher reconstruction error compared to the other frames in $V_i$ and vice versa. Therefore, we calculate the final anomaly score $\mathcal{A}_t$ as:
\begin{equation}
    \mathcal{A}_t = 1 - \mathcal{Q}_t \text{.}
\label{eq:anomalyscore}
\end{equation}
The summary of anomaly score calculation is illustrated in Fig. \ref{fig:overalltraintest}(b). Furthermore, the process of calculating anomaly scores from a test video is described in Algorithm $\ref{algo:test}$.

\begin{algorithm}
\caption{Anomaly score calculation of a test video.}
\label{algo:test}
\begin{algorithmic}[1]
\Require{a test video $V = \{V_1, V_2, \cdots, V_l\}$ of length $l$, trained autoencoder $\mathcal{AE}$, function to measure array length $length$} 
\Ensure{$T$ (sequence length of AE input)}
\State{$I \gets [\;]$}
\State{$\hat{I} \gets [\;]$}
\For{$i = 1$ to $l-T+1$} 
    \State{$X = \{X_1, X_2, \cdots, X_T\}=\{V_i, V_{i+1}, \cdots, V_{i+T-1}\}$}
    \State{$\hat{X}= \{\hat{X}_1, \hat{X}_2, \cdots, \hat{X}_T\} = \mathcal{AE}(X)$}
    \State{$I \gets $ append $X_{1+\left \lfloor{ \frac{T}{2}}\right \rfloor }$} \Comment{middle frame of $X$}
    \State{$\hat{I} \gets $ append $\hat{X}_{1+\left \lfloor{ \frac{T}{2}}\right \rfloor }$} \Comment{middle frame of $\hat{X}$}
\EndFor
\State{$\mathcal{P} \gets [\;]$}
\For{$t = 1$ to $length(I)$} 
    \State{$\mathcal{P} \gets$ append $10 \text{log}_{10}  \frac{M_{\hat{I}_t}^2}{\frac{1}{R} \left \| \hat{I}_t - I_t  \right \|_{F}^{2}}$} \Comment{Eq. \eqref{eq:psnr}}
\EndFor
\For{$t = 1$ to $length(I)$} 
    \State{$\mathcal{Q}_t = \frac{\mathcal{P}_t - \min_t(\mathcal{P}_t)}{\max_t(\mathcal{P}_t)-\min_t(\mathcal{P}_t)}$} \Comment{Eq. \eqref{eq:minmax}}
    \State{$\mathcal{A}_t = 1 - \mathcal{Q}_t$} \Comment{Eq. \eqref{eq:anomalyscore}}
\EndFor
\end{algorithmic}
\end{algorithm}

\section{Experiments}
\label{sec:experiments}

\subsection{Datasets}
We evaluate our approach on three benchmark video anomaly detection datasets, i.e., Ped2 \cite{li2013anomaly}, Avenue \cite{lu2013abnormal}, and ShanghaiTech \cite{luo2017revisit}, which are publicly available. 
We use the standard division of the datasets where the training set contains only normal videos while each video in the test set contains one or more anomalous events.

\noindent\textbf{Ped2.} This dataset contains 16 training and 12 test videos \cite{li2013anomaly}. Normal frames mostly include pedestrians, whereas anomalous frames include bicycles, vehicles, or skateboards.

\noindent\textbf{Avenue.} This dataset consists of 16 training and 21 test videos \cite{lu2013abnormal}. The anomalous events include abnormal objects such as bicycles and abnormal actions of humans such as walking towards unusual directions, throwing stuff, or running.

\noindent\textbf{ShanghaiTech.} This is by far the largest one-class anomaly detection dataset consisting of 330 training and 107 test videos \cite{luo2017revisit}.
The dataset is recorded at 13 different locations with various camera angles and lighting conditions. In total, there are 130 anomalous events in the test videos including fighting, running, and riding bicycle.

\begin{table}[]
\caption{Default hyperparameters used in the experiments.}
\resizebox{\linewidth}{!}{
\begin{tabular}{|cc||cc||ccccc||c||cc|}
\hline
\multicolumn{2}{|c||}{Skip frames}        & \multicolumn{2}{c||}{Repeat frames}   & \multicolumn{5}{c||}{Patch}                                                                                                                      & Fusion & \multicolumn{2}{c|}{Noise}       \\ \hline
\multicolumn{1}{|c|}{$p$}    & $s$           & \multicolumn{1}{c|}{$p$}    & $r$        & \multicolumn{1}{c|}{$p$}    & \multicolumn{1}{c|}{$\alpha$} & \multicolumn{1}{c|}{$\beta$} & \multicolumn{1}{c|}{patching technique} & intruder dataset & $p$      & \multicolumn{1}{c|}{$p$}    & $\sigma$  \\ \hline \hline
\multicolumn{1}{|c|}{0.01} & \{2, 3, 4, 5\} & \multicolumn{1}{c|}{0.01} & \{2, 3\} & \multicolumn{1}{c|}{0.01} & \multicolumn{1}{c|}{0.5}   & \multicolumn{1}{c|}{25}   & \multicolumn{1}{c|}{SmoothMixS \cite{lee2020smoothmix}  }       & CIFAR-100 \cite{krizhevsky2009learning}       & 0.01   & \multicolumn{1}{c|}{0.01} & 0.05 \\ \hline
\end{tabular}
}
\label{tab:defaulthyperparameters}
\end{table}

\subsection{Experimental setup}
\noindent\textbf{Evaluation metric.}
To measure the performance of our approach, we follow the commonly used frame-level area under the ROC curve (AUC) metric \cite{zaheer2020old}. The ROC curve is obtained by plotting false and true positive rates across different anomaly score thresholds. We calculate one ROC curve for the whole test set of a dataset.
Higher AUCs represent more accurate results.


\noindent\textbf{Parameters and implementation details.} 
As an autoencoder model, we adopt a 3D convolution-transposed convolution architecture 
proposed by Gong \etal {} \cite{gong2019memorizing}
which takes an input sequence $X$ of size $16 \times 1 \times 256 \times 256$ and produces its reconstruction of the same size.
The memory mechanism between the encoder and the decoder is removed. Moreover, a Tanh layer is added to the output to limit the output range to $[-1, 1]$.
All the 16 frames of an input and its output sequence are used for calculating the reconstruction loss during training (Eq. \eqref{eq:aereconloss} \& \eqref{eq:pseudoreconloss}).
At test time, 
only the 9th frame is considered for anomaly score calculations (Eq. \eqref{eq:psnr} - \eqref{eq:anomalyscore}). 
The implementation of our models is performed in PyTorch \cite{NEURIPS2019_9015}. The training is carried out using Adam \cite{kingma2014adam} optimizer with a learning rate of $10^{-4}$ and the minibatch size is set to $4$. 
Unless specified otherwise, we use hyperparameters specified in Table \ref{tab:defaulthyperparameters}. 

\subsection{Comparisons with the baseline}
Comparisons of AUC results between the baseline and our method can be seen in the last six rows of Table \ref{tab:sota}. 
The baseline refers to an AE trained without pseudo anomalies 
($p=0$ in Eq. \eqref{eq:inputselection}). Regardless of the pseudo anomaly synthesizing techniques, all of our models successfully outperform the baseline. This demonstrates the importance and the generalization capability of training using pseudo anomalies in order to generate highly discriminative models.

\begin{table}[]
\caption{Comparisons of frame-level AUC performance between our approach and existing SOTA methods on Pedestrian2 (Ped2), Avenue (Ave), and ShanghaiTech (Sh) datasets. Best and second best performances in each category and dataset are marked as bold and underlined.
}
\resizebox{\linewidth}{!}{
\centering
\begin{tabular}[t]{c|l|ccc|}
\hline
\multicolumn{2}{c|}{Methods}                  & Ped2 \cite{li2013anomaly}  & Ave \cite{lu2013abnormal} & Sh \cite{luo2017revisit}     \\ \hline \hline
\parbox[t]{2mm}{\multirow{16}{*}{\rotatebox[origin=c]{90}{Miscellaneous}}}           
  & AbnormalGAN \cite{ravanbakhsh2017abnormal}     & 93.5\%  & -       & -       \\
  & STAN \cite{lee2018stan}                        & 96.5\%  & 87.2\%  & -       \\
  & MC2ST \cite{liu2018classifier}                 & 87.5\%  & 84.4\%  & -       \\
  & Ionescu \etal   \cite{ionescu2019detecting}    & -       & 88.9\%  & -       \\
  & BMAN \cite{lee2019bman}                        & 96.6\%  & \textbf{90.0\%}  & \underline{76.2\%}  \\
  & AMC  \cite{nguyen2019anomaly}                  & 96.2\%  & 86.9\%  & -       \\
  & Vu \etal \cite{vu2019robust}                   & \textbf{99.21\%}  & 71.54\%  & -       \\
  & DeepOC   \cite{wu2019deep}                     & -     & 86.6\%  & -       \\
  & TAM-Net  \cite{ji2020tam}                      & \underline{98.1\%}  & 78.3\%  & -       \\
  & LSA \cite{abati2019latent}                     & 95.4\%  & -       & 72.5\%  \\
  & Ramachandra \etal \cite{ramachandra2020learning} & 94.0\%  & \underline{87.2\%}  & -   \\
  & Tang \etal \cite{tang2020integrating}          & 96.3\%  & 85.1\%  & 73.0\%  \\
  & Wang \etal \cite{wang2020cluster}              & -       & 87.0\%  & \textbf{79.3\%}  \\
  & OGNet \cite{zaheer2020old}                     & \underline{98.1\%}  & -       & -       \\
  & Conv-VRNN \cite{lu2019future}         & 96.06\%  & 85.78\%  & -  \\ 
  & Chang \etal \cite{chang2020clustering}         & 96.5\%  & 86.0\%  & 73.3\%  \\ \hline
  
  \parbox[t]{2mm}{\multirow{9}{*}{\rotatebox[origin=c]{90}{Non deep learning}}}
  & MPPCA \cite{kim2009observe}                        & 69.3\%  & -       & -     \\
  & Mehran \etal \cite{mehran2009abnormal}             & 55.6\%  & -       & -     \\
  & MDT   \cite{mahadevan2010anomaly}                  & 82.9\%  & -       & -     \\
  & Lu \etal \cite{lu2013abnormal}                     & -       & \textbf{80.9\%}  & -     \\
  & AMDN   \cite{xu2017detecting}                      & \underline{90.8\%}  & -       & -     \\
  & Del Giorno \etal \cite{del2016discriminative}      & -       & \underline{78.3\%}  & -     \\
  & LSHF    \cite{zhang2016video}                      & \textbf{91.0\%}  & -       & -     \\
  & Xu \etal \cite{xu2014video} & 88.2\%  & -  & -     \\
  & Ramachandra and Jones \cite{ramachandra2020street} & 88.3\%  & 72.0\%  & -     \\ \hline

\end{tabular}
\begin{tabular}[t]{c|l|ccc|}
\hline
\multicolumn{2}{c|}{Methods}                  & Ped2 \cite{li2013anomaly}  & Ave \cite{lu2013abnormal} & Sh \cite{luo2017revisit}     \\ \hline \hline
\parbox[t]{2mm}{\multirow{6}{*}{\rotatebox[origin=c]{90}{Object-centric}}}   
  & MT-FRCN \cite{hinami2017joint}                     & 92.2\%  & -       & -       \\
  & Ionescu \etal \cite{ionescu2019object} \footnotemark             & 94.3\%  & 87.4\%  & \underline{78.7\%}  \\
  & Doshi and Yilmaz \cite{doshi2020any,doshi2020continual} & \underline{97.8\%}  & 86.4\%  & 71.62\%  \\
  & Sun \etal \cite{sun2020scene}                      & -       & \underline{89.6\% } & 74.7\%  \\
  & VEC \cite{yu2020cloze}                             & 97.3\%  & \underline{89.6\% } & 74.8\%  \\ 
  & Georgescu \etal \cite{georgescu2021background}             & \textbf{98.7\%}  & \textbf{92.3\%}  & \textbf{82.7\%} \\ \hline
\parbox[t]{2mm}{\multirow{4}{*}{\rotatebox[origin=c]{90}{Prediction}}}        
  & Frame-Pred  \cite{liu2018future}          & 95.4\%  & 85.1\%  & 72.8\%  \\
  & Dong \etal \cite{dong2020dual}            & 95.6\%  & 84.9\%  & \underline{73.7\%}  \\
  & Lu \etal \cite{lu2020few}                 & \underline{96.2\%}  & \underline{85.8\%}  & \textbf{77.9\%}  \\
  & MNAD-Prediction \cite{park2020learning}         & \textbf{97.0\%}  & \textbf{88.5\%}  & 70.5\%  \\ \hline
\parbox[t]{2mm}{\multirow{15}{*}{\rotatebox[origin=c]{90}{Reconstruction}}}   
  & AE-Conv2D  \cite{hasan2016learning}          & 90.0\%  & 70.2\%  & 60.85\% \\
  & AE-Conv3D  \cite{zhao2017spatio}             & 91.2\%  & 71.1\%  & -       \\
  & AE-ConvLSTM  \cite{luo2017remembering}       & 88.10\% & 77.00\% & -       \\
  & TSC \cite{luo2017revisit}                    & 91.03\% & 80.56\% & 67.94\% \\
  & StackRNN \cite{luo2017revisit}               & 92.21\% & 81.71\% & 68.00\% \\
  & MemAE \cite{gong2019memorizing}              & 94.1\%  & 83.3\%  & 71.2\%  \\
  & MNAD-Reconstruction \cite{park2020learning}           & 90.2\%  & 82.8\%  & 69.8\%  \\ 
  & LNTRA-Skip frames \cite{astrid2021learning}      & 96.50\% & 84.67\% & \textbf{75.97\%} \\
  & LNTRA-Patch \cite{astrid2021learning}     &  94.77\% & 84.91\% & 72.46\% \\ \cdashline{2-5} 
  & Baseline              & 92.49\%  & 81.47\% & 71.28\% \\
  & PseudoBound-Skip frames      & \textbf{98.44\%} & \textbf{87.10\%} & \underline{73.66\%} \\
  & PseudoBound-Repeat frames      & 93.69\% & 81.87\% & 72.58\% \\
  & PseudoBound-Patch      & 95.33\% & \underline{85.36\%} & 72.77\% \\
  & PseudoBound-Fusion      &  94.16\% & 82.79\% & 71.52\% \\ 
  & PseudoBound-Noise      &  \underline{97.78\%} & 82.11\% & 72.02\% \\\hline
\end{tabular}
}
\footnotemark[\value{footnote}]\footnotesize{Micro AUC reported in \cite{georgescu2021background}}
\label{tab:sota}
\end{table}

Moreover, the capability of the models in producing discriminative reconstructions can also be observed in Fig. \ref{fig:qualitative}. 
The reconstructions and the corresponding error heatmaps of our methods can be seen in Fig. \ref{fig:qualitative}(b)-(f), (k)-(o), and (s)-(w). Comparing with the results produced by the baseline in Fig. \ref{fig:qualitative}(a), (j), and (r), our models trained using any type of the pseudo anomaly successfully distort the reconstructions of abnormal regions better than the baseline.

\subsection{Comparisons with other methods}
\begin{figure}
\begin{center}
\includegraphics[width=\linewidth]{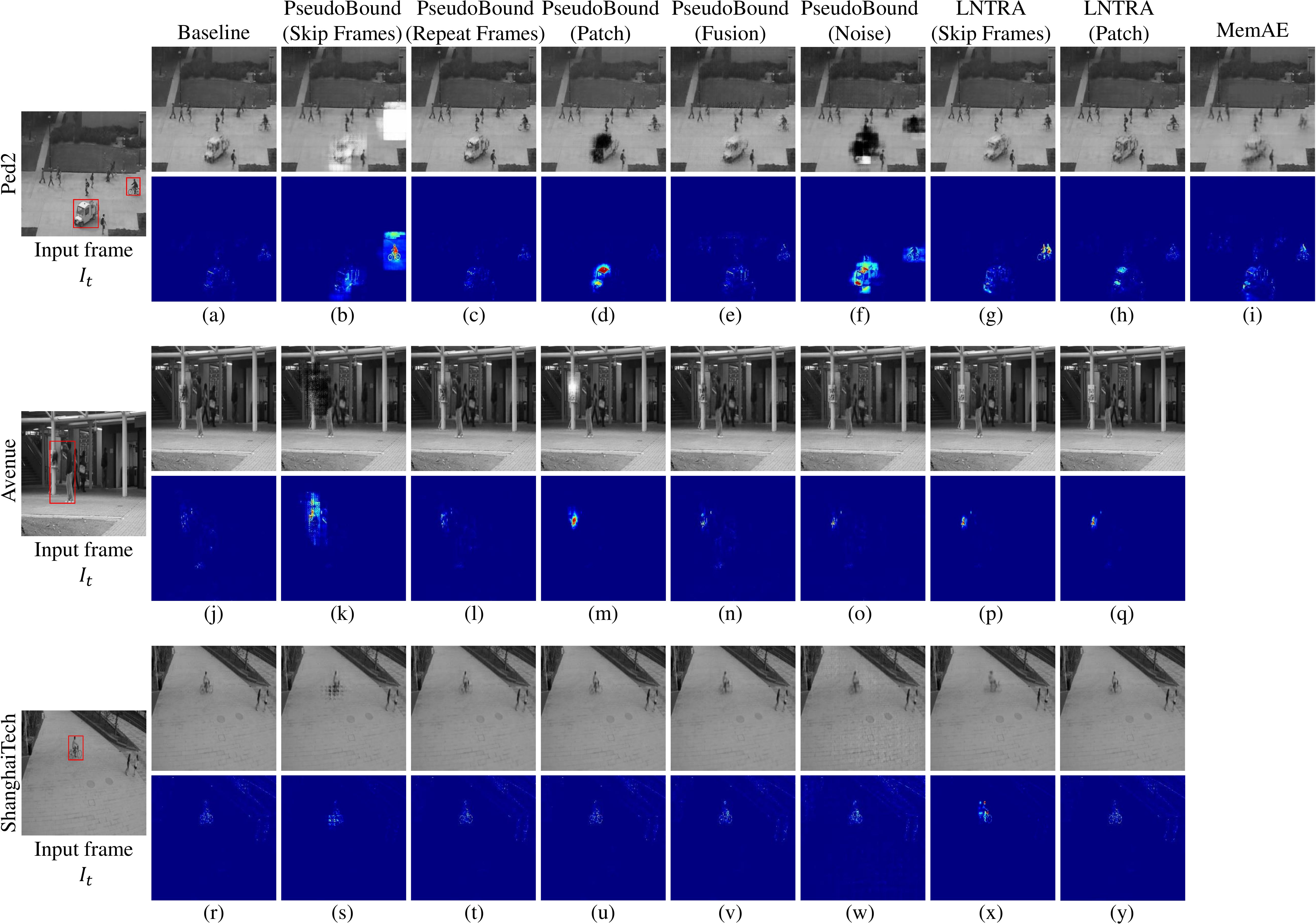}
\end{center}
\caption{Comparisons of the reconstructed frames (top row) and their corresponding reconstruction error heatmaps (bottom row) between the baseline, our proposed PseudoBound using different types of pseudo anomaly, LNTRA \cite{astrid2021learning}, and MemAE \cite{gong2019memorizing} on the test frames of three benchmark datasets. The heatmaps are generated by min-max normalizing the squared error of each pixel between the reconstruction and the input frame. 
Anomalous regions are marked with red boxes.}
\label{fig:qualitative}
\end{figure}

Table \ref{tab:sota} shows the frame-level AUC comparisons of our PseudoBound with several existing state-of-the-art (SOTA) approaches on Ped2 \cite{li2013anomaly}, Avenue \cite{lu2013abnormal}, and ShanghaiTech \cite{luo2017revisit} datasets. 
Similar to \cite{astrid2021learning}, we classify the existing SOTA approaches into five categories: 1) non-deep learning methods, 2) object-centric approaches that apply object detection before anomaly detection to focus only on the detected objects, 3) prediction-based approaches that predict future frames to detect anomalies, 4) reconstruction-based methods that use reconstruction to detect anomalies, and 5) miscellaneous approaches which do not belong to any of the categories or employing a combination of the categories. Furthermore, the qualitative results of several SOTA approaches \cite{astrid2021learning,gong2019memorizing} are also provided in Fig. \ref{fig:qualitative}(g)-(i), (p)-(q), and (x)-(y).

\subsubsection{Comparison with other reconstruction-based methods that use pseudo anomalies}
\label{subsubsec:comparewithlntra}
Another way to train a reconstruction-based model using pseudo anomalies has also been proposed in \cite{astrid2021learning}. In the case of pseudo anomaly input, LNTRA \cite{astrid2021learning} is trained to minimize the reconstruction loss between the model's reconstruction and the corresponding normal data that is used to synthesize the pseudo anomaly input. In contrast, our method maximizes the reconstruction loss. As seen in Table \ref{tab:sota}, by comparing the results between the two losses using the same type of pseudo anomalies, i.e., skip frames and patch, on average of three datasets, PseudoBound-Skip frames achieves a better performance of $86.40\%$ compared to LNTRA-Skip frames with $85.71\%$. Similarly, PseudoBound-Patch with $84.49\%$ compared to LNTRA-Patch with $84.05\%$. 

However, interestingly, using skipping frames pseudo anomalies, the models trained using the loss in \cite{astrid2021learning} tend to be more effective in a more complex environments such as ShanghaiTech. This may be caused by the difference of the innate properties of each loss. In our proposed PseudoBound, maximizing the loss in the case of pseudo anomalies enforces the model to distort the reconstructions as much as possible. This can lead to a more discriminative reconstructions compared to the loss used in \cite{astrid2021learning}, hence the better performance compared to \cite{astrid2021learning}. However, the problem is that there are also normal regions in the pseudo anomalies, e.g., the background. Moreover, pseudo anomalies are not real anomalies and have some level of similarity to the normal data. Maximizing the reconstruction loss as it is can sometimes be harmful to the model as the normal parts can also be distorted. In such case, the loss in \cite{astrid2021learning} is more forgiving as the training objective is to generate normal frames even if the input is pseudo anomalies. This property may contribute to the better performance of LNTRA-skip frames \cite{astrid2021learning} in the more complex dataset.

The difference between each loss properties can also be clearly observed qualitatively. 
PseudoBound models try to distort the anomalous regions directly as can be seen in, for example, Fig. \ref{fig:qualitative}(b), (d), (k), (m), and (s).
On the other hand, the reconstructions in LNTRA \cite{astrid2021learning} are not so straightforwardly distorted, for example in Fig. \ref{fig:qualitative}(g), (p), (q), and (x). Instead, LNTRA models try to reconstruct normal only which indirectly leads to the high reconstruction error. Comparing the reconstructions of normal regions (e.g., the normal pedestrians), for example between Fig. \ref{fig:qualitative}(b) of PseudoBound and (g) of LNTRA, the normal regions in the PseudoBound (Fig. \ref{fig:qualitative}(b)) looks slightly more distorted compared to the normal regions produced in LNTRA \cite{astrid2021learning} (Fig. \ref{fig:qualitative}(g)). However, since the large distortion of anomalous regions outweighs the slight distortion of the normal regions, PseudoBound models can overall outperform the models proposed in \cite{astrid2021learning}.

As for the reason why PseudoBound models with some types of pseudo anomalies tend to be inferior compared to the other types and other reconstruction-based approaches, please refer to the discussions on Section \ref{subsec:discussions}-Why are some pseudo anomaly types more inferior than the others?

\subsubsection{Comparison with reconstruction-based methods that use memory mechanism}
As seen in Table \ref{tab:sota}, using average AUC in three datasets, most of the PseudoBound models achieve superior performance compared to the memory-based networks, such as MemAE \cite{gong2019memorizing} and MNAD-Reconstruction \cite{park2020learning}. The only exception is in MemAE achieving a slightly better average AUC of $82.87\%$ over the three datasets than PseudoBound-Repeat frames with $82.71\%$, which can be attributed to the choice of pseudo anomalies. This demonstrates the superiority of our methods in limiting the reconstruction capability of AE compared to the memory-based network. As also discussed in Section \ref{sec:introduction} and \ref{sec:relatedworks}, the memory mechanisms may severely distort the normal regions as well, which reduces the discrimination capability of the model. This property can also be observed in Fig. \ref{fig:qualitative}(i), where MemAE \cite{gong2019memorizing} poorly reconstructs the normal pedestrians.

\subsubsection{Comparison with non-reconstruction methods} Compared to the techniques in other categories, PseudoBound demonstrates a comparable performance.
In the object-centric category, most approaches \cite{ionescu2019object,doshi2020any,doshi2020continual,sun2020scene,yu2020cloze,georgescu2021background} require object detectors, which limits their anomaly detectors only to the set of predefined object categories. These methods also enjoy the benefits of the specialized object-centric training, e.g. background removal, which is reflected by the remarkable performance of \cite{ionescu2019object}.
On the other hand, our work is not limited by the predefined object categories thus can be applied generically to any types of anomalies, whether they are object or non-object categories.

On the other hand, several methods in miscellaneous are particularly designed with complex components. For example, BMAN {\cite{lee2019bman}} utilizes attention, adversarial training, bidirectional LSTM, and multi-scale setup all in one model. Moreover, the anomaly detection approach by Vu \textit{et al.} {\cite{vu2019robust}} requires a pair of cGANs (conditional Generative Adversarial Networks), optical flow input, and multiple denoising AEs. In contrast, our method provides comparable performance without any bells and whistles, i.e. only requires an autoencoder with a sequence of images as input. Moreover, our mechanism does not add any computational cost on top of the baseline at test time.
It may be noted that the purpose of our experiments is not to pursue the best performance on certain categories but to exhibit the possibility of our approach being a generic solution towards training one-class video anomaly detectors that can be directly incorporated to several existing AE-based approaches \cite{hasan2016learning,zhao2017spatio,luo2017remembering,luo2017revisit,zaheer2020old}.

\subsection{Additional discussions}
\label{subsec:discussions}
In this section, we discuss several topics which have not been covered, such as whether we need prior knowledge to design pseudo anomaly and robustness towards different hyperparameters values.

\noindent\textbf{Why are some pseudo anomaly types more inferior than the others?} Synthesizing pseudo anomalies using skipping frames relies on the assumption that anomalous behaviors are largely related to the fast movements. As such behaviors exist abundantly in the test set, our model trained using skip frame pseudo anomalies works the best compared to the other types of pseudo anomalies, as seen in Table \ref{tab:sota}. To observe whether prior knowledge is required to design pseudo anomalies, we propose repeating frames as the opposite of skipping frames. Moreover, we also propose adding noise which does not require prior knowledge. 
Patch and fusion pseudo anomalies are based on general prior knowledge about unusual shapes of anomalous objects. However, such anomalous patterns rarely appear in the test dataset, especially the fusion type that produces blurry objects.  Therefore, considering the performance of skipping frames outperforms the others, prior knowledge on the test cases is good to have.
However, since all the pseudo anomalies proposed in this paper achieve superior performance compared to the baseline, prior knowledge is not the only way and thus not necessarily required to limit the reconstruction capability of AEs.

\begin{figure}
\begin{center}
\includegraphics[width=\linewidth]{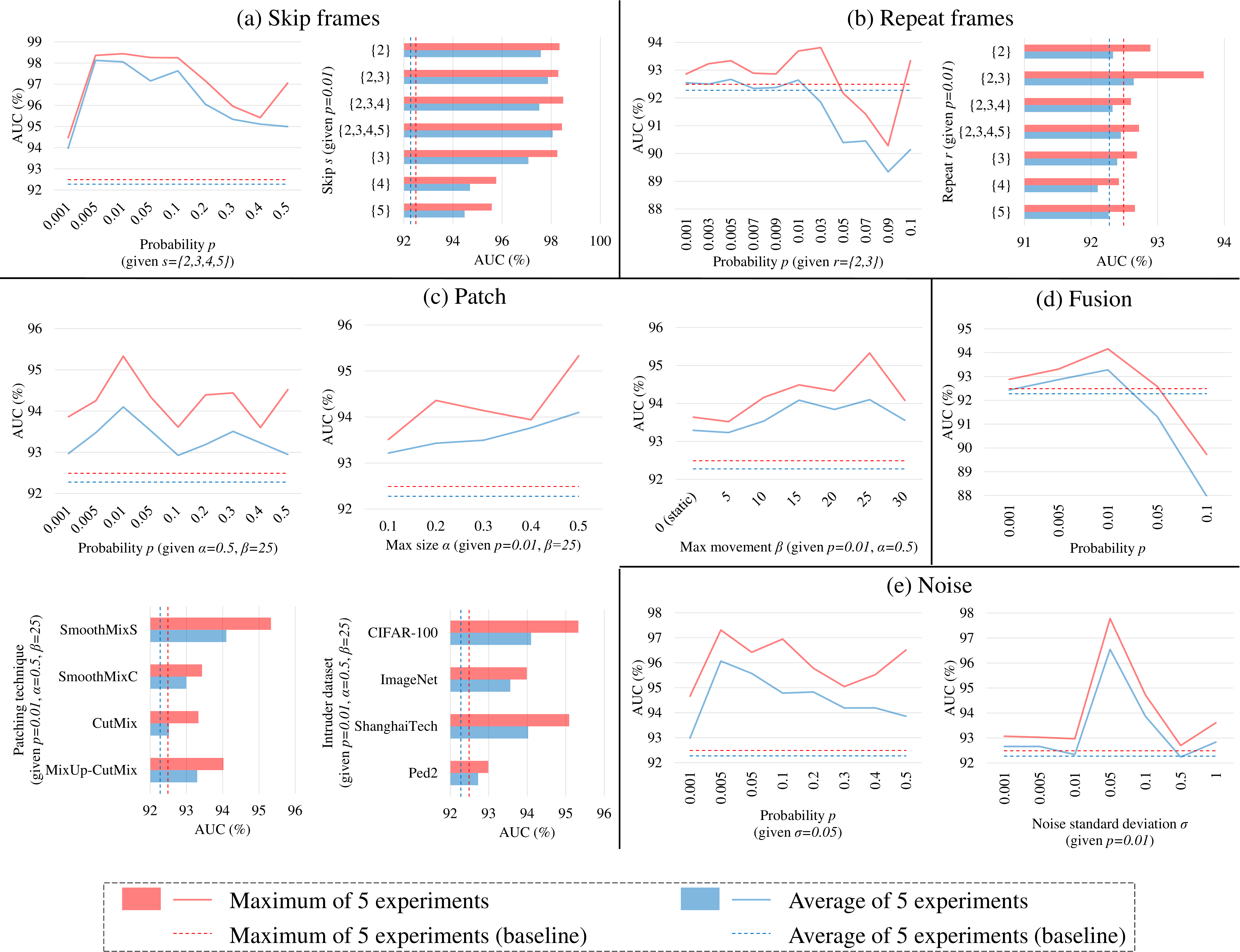}
\end{center}
\caption{Evaluations on different hyperparameter values in each pseudo anomaly synthesizing technique. To limit the span of experiments, we evaluate only in Ped2. }
\label{fig:hyperparametereval}
\end{figure}

\noindent\textbf{How many pseudo anomalies are required for the training?} The number of pseudo anomalies is controlled by a hyperparameter probability $p$. As seen in Table \ref{tab:defaulthyperparameters}, we basically need only a small amount of pseudo anomalies, such as $1\%$ when $p=0.01$. Rather, we observe that too much pseudo anomalies can degrade the model performance, as seen in the experiments using higher $p$ values in Fig. \ref{fig:hyperparametereval}. It may be because pseudo anomalies are not real anomalies such that they have normal parts as well and some level of similarity to the normal data. 

\noindent\textbf{How robust each of pseudo anomaly synthesizing technique is?} Fig. \ref{fig:hyperparametereval} shows the experiments using different $s$, $r$, $\alpha$, $\beta$, patching technique, intruding dataset, and $\sigma$ values. In general, our method is robust to a wide range of values, thus successfully outperforms the baseline.

\begin{table}[]
\caption{Frame-level AUC on Ped2 using different combinations of pseudo anomaly types. \# denotes the number pseudo anomaly types. Note that, model trained using 0 number of pseudo anomaly type is the baseline model. The best AUC for each number of pseudo anomaly types is marked as bold.
}
\resizebox{\linewidth}{!}{
\centering
\begin{tabular}{|c||c|c|c|c|c||c|}
\hline
\# & Skip & Patch & Fusion & Noise & Repeat & AUC   \\ \hline\hline
0                             &            &       &         &       &               & \textbf{92.49\%} \\ \hline\hline
\multirow{5}{*}{1}            & \checkmark          &       &         &       &               & \textbf{98.44\%} \\ \cline{2-7} 
                              &            & \checkmark     &         &       &               & 95.33\% \\ \cline{2-7} 
                              &            &       & \checkmark       &       &               & 94.16\% \\ \cline{2-7} 
                              &            &       &         & \checkmark     &               & 97.78\% \\ \cline{2-7} 
                              &            &       &         &       & \checkmark             & 93.69\% \\ \hline\hline
\multirow{10}{*}{2}           & \checkmark          & \checkmark     &         &       &               & 98.33\% \\ \cline{2-7} 
                              & \checkmark          &       & \checkmark       &       &               & 98.07\% \\ \cline{2-7} 
                              & \checkmark          &       &         & \checkmark     &               & \textbf{98.70\%} \\ \cline{2-7} 
                              & \checkmark          &       &         &       & \checkmark             & 98.42\% \\ \cline{2-7} 
                              &            & \checkmark     & \checkmark       &       &               & 94.56\% \\ \cline{2-7} 
                              &            & \checkmark     &         & \checkmark     &               & 94.74\% \\ \cline{2-7} 
                              &            & \checkmark     &         &       & \checkmark             & 95.21\% \\ \cline{2-7} 
                              &            &       & \checkmark       & \checkmark     &               & 95.86\% \\ \cline{2-7} 
                              &            &       & \checkmark       &       & \checkmark             & 93.40\% \\ \cline{2-7} 
                              &            &       &         & \checkmark     & \checkmark             & 98.67\% \\ \hline

\end{tabular}
\begin{tabular}{|c||c|c|c|c|c||c|}
\hline
\# & Skip & Patch & Fusion & Noise & Repeat & AUC   \\ \hline\hline
\multirow{10}{*}{3}           &            &       & \checkmark       & \checkmark     & \checkmark             & 96.52\% \\ \cline{2-7} 
                              &            & \checkmark     &         & \checkmark     & \checkmark             & 96.21\% \\ \cline{2-7} 
                              &            & \checkmark     & \checkmark       &       & \checkmark             & 93.82\% \\ \cline{2-7} 
                              &            & \checkmark     & \checkmark       & \checkmark     &               & 96.35\% \\ \cline{2-7} 
                              & \checkmark          &       &         & \checkmark     & \checkmark             & 98.11\% \\ \cline{2-7} 
                              & \checkmark          &       & \checkmark       &       & \checkmark             & 97.83\% \\ \cline{2-7} 
                              & \checkmark          &       & \checkmark       & \checkmark     &               & 98.55\% \\ \cline{2-7} 
                              & \checkmark          & \checkmark     &         &       & \checkmark             & 98.43\% \\ \cline{2-7} 
                              & \checkmark          & \checkmark     &         & \checkmark     &               & \textbf{98.78\%} \\ \cline{2-7} 
                              & \checkmark          & \checkmark     & \checkmark       &       &               & 98.48\% \\ \hline\hline
\multirow{5}{*}{4}            &            & \checkmark     & \checkmark       & \checkmark     & \checkmark             & 93.66\% \\ \cline{2-7} 
                              & \checkmark          &       & \checkmark       & \checkmark     & \checkmark             & 98.70\% \\ \cline{2-7} 
                              & \checkmark          & \checkmark     &         & \checkmark     & \checkmark             & 98.72\% \\ \cline{2-7} 
                              & \checkmark          & \checkmark     & \checkmark       &       & \checkmark             & 98.46\% \\ \cline{2-7} 
                              & \checkmark          & \checkmark     & \checkmark       & \checkmark     &               & \textbf{98.77\%} \\ \hline\hline
5                             & \checkmark          & \checkmark     & \checkmark       & \checkmark     & \checkmark             & \textbf{98.97\%} \\ \hline

\end{tabular}
}
\label{tab:pseudoanomalycombination}
\end{table}

\noindent\textbf{Combining multiple types of pseudo anomaly.}
To make anomaly detector more robust towards different types of anomaly, using multiple types of pseudo anomaly to train a model can be done in a similar way as augmentation techniques in image classification. We conduct experiments with all of the possible combinations using the proposed five types of pseudo anomalies on Ped2 dataset. In this series of experiments, using the default hyperparameters in Table {\ref{tab:defaulthyperparameters}}, the probability of multiple types of pseudo anomalies is added. For example, if skip frames and patch are used, the probability of skip frame, patch, and normal is $0.01$, $0.01$, and $0.98$. If all the five types are used, the probability of using normal data is $0.95$. The results on Ped2 can be seen in Table {\ref{tab:pseudoanomalycombination}}. As seen, using multiple types of pseudo anomalies can make our model more robust, hence improving the AUC performance.

\noindent\textbf{Anomaly score: PSNR vs MSE.}
In order to calculate PSNR (Eq. {\eqref{eq:psnr}}), Mean Squared Error (MSE) is used. Therefore, PSNR theoretically has similar purpose to MSE, except that it has an opposite trend and a non-linearity using log. We re-evaluate our trained model in Table {\ref{tab:sota}} using MSE to calculate $\mathcal{P}_t$ followed by normalization in Eq. {\eqref{eq:minmax}} to get the final anomaly score. Note that we do not flip the anomaly score in Eq. {\eqref{eq:anomalyscore}} as lower values of MSE already represent lower anomaly scores and vice versa. The results can be seen in Table  {\ref{tab:psnrvsmse}}. As seen, we find experimentally that PSNR is a better choice than MSE in our method.

\begin{table}[]
\caption{AUC comparisons for PSNR and MSE. The better performance between the two methods is marked as bold.} 
\resizebox{\linewidth}{!}{
\begin{tabular}{|l||cc||cc||cc|}
\hline
\multirow{2}{*}{}         & \multicolumn{2}{c||}{Ped2}                            & \multicolumn{2}{c||}{Avenue}                 & \multicolumn{2}{c|}{ShanghaiTech}           \\ \cline{2-7} 
                          & \multicolumn{1}{c|}{PSNR}           & MSE            & \multicolumn{1}{c|}{PSNR}           & MSE   & \multicolumn{1}{c|}{PSNR}           & MSE   \\ \hline\hline
PseudoBound-Skip frames   & \multicolumn{1}{c|}{\textbf{98.44\%}} & 98.39\%          & \multicolumn{1}{c|}{\textbf{87.10\%}} & 84.37\% & \multicolumn{1}{c|}{\textbf{73.66\%}} & 72.31\% \\ \hline
PseudoBound-Repeat frames & \multicolumn{1}{c|}{93.69\%}          & \textbf{94.20\%} & \multicolumn{1}{c|}{\textbf{81.87\%}} & 80.71\% & \multicolumn{1}{c|}{\textbf{72.58\%}} & 71.64\% \\ \hline
PseudoBound-Patch         & \multicolumn{1}{c|}{\textbf{95.33\%}} & 95.15\%          & \multicolumn{1}{c|}{\textbf{85.36\%}} & 78.26\% & \multicolumn{1}{c|}{\textbf{72.77\%}} & 72.03\% \\ \hline
PseudoBound-Fusion        & \multicolumn{1}{c|}{94.16\%}          & \textbf{94.34\%} & \multicolumn{1}{c|}{\textbf{82.79\%}} & 81.60\%  & \multicolumn{1}{c|}{\textbf{71.52\%}} & 70.97\% \\ \hline
PseudoBound-Noise         & \multicolumn{1}{c|}{\textbf{97.78\%}} & 97.31\%          & \multicolumn{1}{c|}{\textbf{82.11\%}} & 79.63\% & \multicolumn{1}{c|}{\textbf{72.02\%}} & 71.69\% \\ \hline
\end{tabular}
}
\label{tab:psnrvsmse}
\end{table}

\noindent\textbf{Computational cost.} Our proposed method of utilizing pseudo anomalies to limit reconstruction capability of AE does not add any computational cost to the baseline during test time. The test mechanism of our PseudoBound models is same as the baseline model, i.e., we feed the test sequence to the AE as it is and calculate anomaly score from the reconstruction loss (Section \ref{subsec:anomalyscore}). On an NVIDIA GeForce Titan Xp, our models performs inference at 40 frames per second.

\section{Conclusions}
In this paper, we propose a training mechanism to limit the reconstruction capability of AE on anomalous data. In addition to enforcing low reconstruction error on normal data, the training mechanism also aims high reconstruction error for anomalous cases with help of pseudo anomalies synthesized via five different ways.
Our experiments using various synthesizing methods in three benchmark datasets demonstrate the robustness of our training mechanism in limiting reconstruction capability of AE. With the effectiveness of pseudo anomaly based training, this work enlightens and emphasizes on the importance of pseudo anomalies in limiting reconstruction capability of AE, which may encourage the community to develop more generic and effective pseudo anomalies for enhanced one-class classifiers. Future directions include high-level pseudo anomalies in feature space and pseudo anomalies that can effectively work not only on video data, but also other types of data, e.g. images, network signals, and sensor readings.  In addition, finding types of pseudo anomaly that are suitable towards different types of architectures, e.g., prediction and object-centric based methods, may also be an interesting topic to explore in the future.


\section*{Acknowledgments} 
\noindent This work was supported by Institute of Information \& communications Technology Planning \& Evaluation (IITP) grant funded by the Korea government (MSIT) (No. 2022-0-00951, Development of Uncertainty-Aware Agents Learning by Asking Questions).


\bibliography{mybibfile}

\section*{Authors} 

\begin{wrapfigure}{l}{25mm} 
    \includegraphics[width=1in,height=1.25in,clip,keepaspectratio]{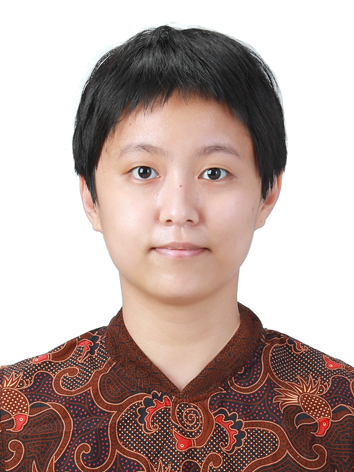}
  \end{wrapfigure}\par
  \textbf{Marcella Astrid} received her BEng in computer engineering from the Multimedia Nusantara University, Tangerang, Indonesia, in 2015, and the MEng in Computer Software from the University of Science and Technology (UST), Daejeon, Korea, in 2017. At the same university, she received her PhD degree in Artificial Intelligence in February 2023. She starts her position as a postdoctoral researcher at the University of Luxembourg from April 2023. Her recent interests include anomaly detection, data augmentation, out of distribution data detection, and computer vision.\par
  
\begin{wrapfigure}{l}{25mm} 
  \includegraphics[width=1in,height=1.25in,clip,keepaspectratio]{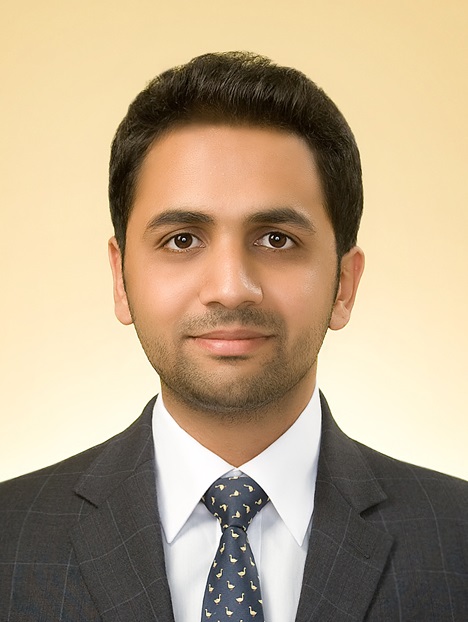}
  \end{wrapfigure}\par
  \textbf{Muhammad Zaigham Zaheer} is currently associated with Mohamed bin Zayed University of Artificial Intelligence (MBZUAI) as a Research Fellow. Previously, he has worked with the Electronics and Telecommunications Research Institute (ETRI), Korea, as a post-doc researcher. Previously, he received his PhD degree from University of Science and Technology, Daejeon, Korea, in 2022, MS degree from Chonnam National University, Gwangju, Korea, in 2017, and undergraduate degree from Pakistan Institute of Engineering and Applied Sciences, Islamabad, Pakistan, in 2012. His current research interests include computer vision, anomaly detection in images/videos, and semi-supervised/self-supervised/unsupervised learning.\par

\begin{wrapfigure}{l}{25mm} 
  \includegraphics[width=1in,height=1.25in,clip,keepaspectratio]{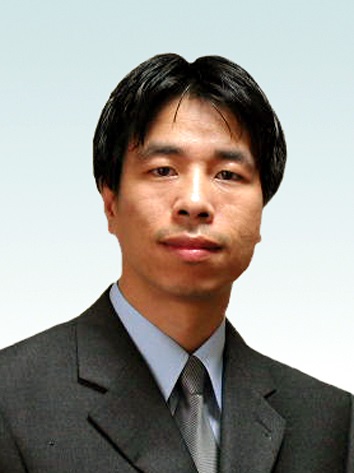}
  \end{wrapfigure}\par
  \textbf{Seung-Ik Lee} received his BS, MS, and PhD degrees in computer science from Yonsei University, Seoul, Korea, in 1995, 1997 and 2001, respectively. He is currently working for the Electronics and Telecommunications Research Institute, Daejeon, Korea. Since 2005, he has been with the Department of Artificial Intelligence, University of Science and Technology, Daejeon, Korea, where he is a professor. His research interests include computer vision, anomaly detection, object detection, open-set learning, out of distribution data detection, domain adaptation, learning with limited data, deep learning, and reinforcement learning.\par

\end{document}